\documentclass[sigconf]{acmart}

\AtBeginDocument{%
  }

\setcopyright{acmcopyright}
\copyrightyear{2023}
\acmYear{2023}
\setcopyright{acmlicensed}
\acmConference[MM '23] {Proceedings of the 31st ACM International Conference on Multimedia}{October 29--November 3, 2023}{Ottawa, ON, Canada}
\acmBooktitle{Proceedings of the 31st ACM International Conference on Multimedia (MM '23), October 29--November 3, 2023, Ottawa, ON, Canada}
\acmPrice{15.00}
\acmDOI{10.1145/3581783.3612404}
\acmISBN{979-8-4007-0108-5/23/10}

\acmSubmissionID{3055}

\usepackage{xspace}
\usepackage{subcaption}
\usepackage{siunitx}
\usepackage{cleveref}
\usepackage{wrapfig}
\usepackage{stmaryrd}
\usepackage{amsfonts}
\usepackage{bm}
\usepackage{amsmath}
\usepackage{multirow}
\usepackage{graphicx}

\usepackage{multirow}
\usepackage{array}

\begin{document}

\title{Deep Image Harmonization in Dual Color Spaces}

\author{Linfeng Tan}
\orcid{0009-0003-3376-0989}
\affiliation{%
  \institution{MoE Key Lab of Artificial Intelligence,}
  \institution{Shanghai Jiao Tong University}
  \country{China}
}
\email{tanlinfeng@sjtu.edu.cn}

\author{Jiangtong Li}
\orcid{0000-0003-3873-4053}
\affiliation{%
   \institution{MoE Key Lab of Artificial Intelligence,}
  \institution{Shanghai Jiao Tong University}
  \country{China}
}
\email{keep_moving-Lee@sjtu.edu.cn}

\author{Li Niu}
\orcid{0000-0003-1970-8634}
\authornote{Corresponding authors.}
\affiliation{%
   \institution{MoE Key Lab of Artificial Intelligence,}
  \institution{Shanghai Jiao Tong University}
  \country{China}
}
\email{ustcnewly@sjtu.edu.cn}

\author{Liqing Zhang}
\authornotemark[1]
\orcid{0000-0001-7597-8503}
\affiliation{%
   \institution{MoE Key Lab of Artificial Intelligence,}
  \institution{Shanghai Jiao Tong University}
  \country{China}
}
\email{zhang-lq@cs.sjtu.edu.cn}

\begin{abstract}

Image harmonization is an essential step in image composition that adjusts the appearance of composite foreground to address the inconsistency between foreground and background. 
Existing methods primarily operate in correlated $RGB$ color space, leading to entangled features and limited representation ability. 
In contrast, decorrelated color space (\emph{e.g.}, $Lab$) has decorrelated channels that provide disentangled color and illumination statistics. 
In this paper, we explore image harmonization in dual color spaces, which supplements entangled $RGB$ features with disentangled $L$, $a$, $b$ features to alleviate the workload in harmonization process. 
The network comprises a $RGB$ harmonization backbone, an $Lab$ encoding module, and an $Lab$ control module. 
The backbone is a U-Net network translating composite image to harmonized image. Three encoders in $Lab$ encoding module extract three control codes independently from $L$, $a$, $b$ channels, which are used to manipulate the decoder features in harmonization backbone via $Lab$ control module. 
Our code and model are available at \href{https://github.com/bcmi/DucoNet-Image-Harmonization}{https://github.com/bcmi/DucoNet-Image-Harmonization}.

\end{abstract}



\maketitle

\section{Introduction}\label{section:Introduction}

\begin{figure}[ht]
  \centering
  \includegraphics[width=0.95\linewidth]{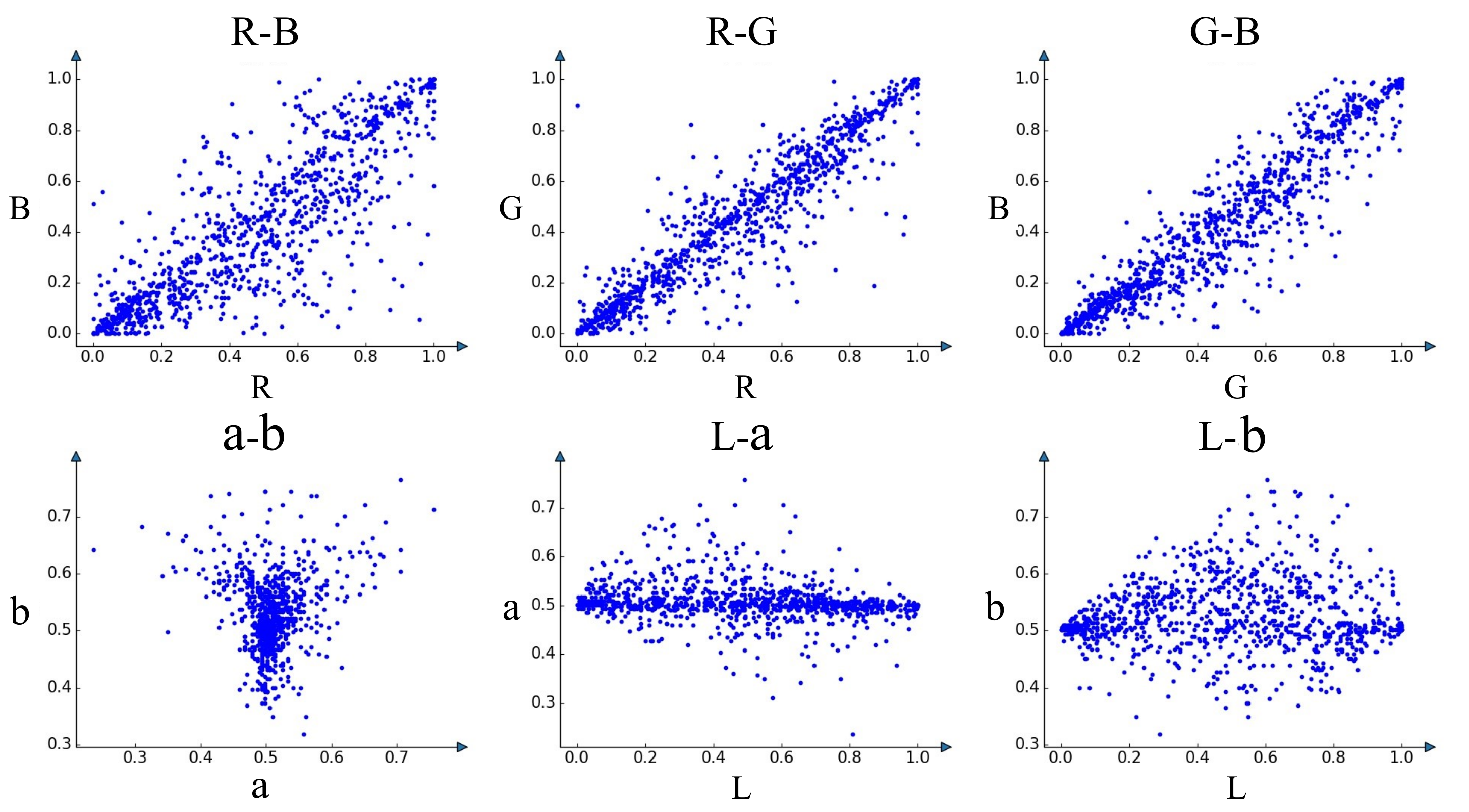}
  \caption{
    We randomly sample 1000 pixels from 100 real images in iHarmony4~\cite{dovenet} and plot the correlation between every two channels in \textit{RGB} (\emph{resp.}, \textit{Lab}) color space in the top (\emph{resp.}, bottom) row. It can be seen that \textit{RGB} channels have strong positive correlations, while \textit{Lab} channels are decorrelated. 
  }
  \label{fig:example}
\end{figure}

Image composition~\cite{niu2021making} targets at generating a composite image by merging foreground and background. Nevertheless, the foreground and background in the obtained composite image might have appearance discrepancy, which is caused by different lighting, climate, and capture devices between foreground and background. To tackle this challenge, image harmonization~\cite{sunkavalli2010multi, xue2012understanding, dovenet, issam, CDTNet} modifies the foreground appearance to ensure its compatibility with the background.
Early traditional image harmonization methods~\cite{sunkavalli2010multi, xue2012understanding, song2020illumination, lalonde2007using} are often designed based on low-level color and illumination statistics. However, with the rapid advance of deep learning techniques, deep image harmonization methods~\cite{dovenet, issam, CDTNet, harmonizer} have become dominant and achieved impressive results.

Existing deep image harmonization methods have been developed from different aspects (\emph{e.g.}, attention mechanism, domain/style transfer, Retinex theory, color transfer) to address the appearance mismatch between foreground and background.
In detail, some works~\cite{ssam, feature_mod} explored attention mechanism to adjust the foreground features more effectively. 
Besides, some works~\cite{dovenet, bargainnet} approached image harmonization as the translation from foreground domain to background domain with additional loss to guide the domain transfer.
Moreover, some works~\cite{intrinsic, IHT} introduced Retinex~\cite{land1971lightness} theory to image harmonization tasks by decoupling an image into reflectance and illumination. 
Recently, some works~\cite{harmonizer, CDTNet} considered the balance between effectiveness and efficiency, and solved image harmonization in the form of color transfer.
Despite the success achieved by existing methods, they mainly operate in $RGB$ color space to extract and adjust features.
However,  $RGB$ color space is a correlated color space and the entangled $RGB$ features may increase the workload of existing harmonization methods.

As known to all, an image can be represented in various color spaces, such as $RGB$, $XYZ$, or $Lab$. 
These color spaces can be categorized into two groups: correlated color spaces and decorrelated color spaces. 
In correlated color spaces (\emph{e.g.}, $RGB$, \textit{XYZ}),  different channels are strongly correlated and tend to change simultaneously. 
In contrast, in decorrelated color spaces (\emph{e.g.}, \textit{YUV}, $Lab$), different channels are decorrelated.
By taking $Lab$ as an example decorrelated color space, \textit{L} represents lightness, \textit{a} represents the spectrum from green to red, and \textit{b} represents the spectrum from blue to yellow. 
In \Cref{fig:example}, we plot the correlation between every two channels in $RGB$ (\emph{resp.}, $Lab$) color spaces in the top (\emph{resp.}, bottom) row. It can be observed that  \textit{RG}, \textit{RB}, and \textit{GB} in $RGB$ color space exhibit strong positive correlations, while \textit{La}, \textit{Lb}, and \textit{ab} in $Lab$ color space are decorrelated.
Considering the correlation within the $RGB$ color space, the extracted $RGB$ features may not effectively disentangle the independent factors of color and illumination statistics, which potentially complicates the harmonization process~\cite{dovenet,issam,rainnet,ssam}. 
However, the decorrelated~$Lab$ color space contains decorrelated factors (\emph{i.e.}, lightness, orthogonal colors) in three channels, serving as a valuable complement to the entangled features extracted from $RGB$ color space. 
Moreover, recent studies~\cite{liang2022inharmonious, wu2022inharmonious} on inharmonious region localization have revealed that the decorrelated color space can help identify the inharmonious region, which also motivates us to explore image harmonization in the decorrelated color space.

Our primary insight for image harmonization is to alleviate the workload of harmonization process by supplementing the entangled $RGB$ features with the disentangled $L$, $a$, $b$ features. 
To this end, we propose a novel image harmonization network in \textbf{Du}al \textbf{Co}lor Spaces (\textbf{DucoNet}). 
Our DucoNet comprises a $RGB$ harmonization backbone, an $Lab$ encoding module, and an $Lab$ control module. 
The harmonization backbone is a U-Net network responsible for harmonizing the input composite image in the $RGB$ color space. In detail, the backbone takes in the $RGB$ channels and the foreground mask, producing the $RGB$ channels of the harmonized image. 
The $Lab$ encoding module consists of three encoders to extract the $L$, $a$, $b$ control codes from $L$, $a$, $b$ channels of the composite image independently. 
The $Lab$ control module interacts with the harmonization backbone to adjust the decoder features with $L$, $a$, $b$ control codes. 
Each control code adjusts the decoder features in multiple decoder layers of the harmonization backbone. Specifically, each control code is used to generate dynamic convolution kernels~\cite{styleganv2}, which are applied to the foreground region in the decoder feature maps. The decoder feature maps manipulated using three control codes are fused to produce the harmonized image. 
Considering that $L$, $a$, $b$ channels may contribute differently to various images or even various pixels, we tend to learn pixel-wise weights for three channels when fusing the decoder feature maps manipulated using three control codes, which could also provide hints for the contributions of $L$, $a$, $b$ channels when harmonizing a specific image. 

The effectiveness of our DucoNet is verified through extensive experiments of low/high-resolution harmonization on the benchmark dataset iHarmony4~\cite{dovenet} and real composite images.
Our contribution can be summarized as follows: 1) To the best of our knowledge, we are the first to investigate image harmonization in both correlated and decorrelated color spaces. 2) We propose a novel image harmonization network in Dual Color Spaces (DucoNet) with  $Lab$ encoding module and control module, which supplements entangled $RGB$ features with disentangled $L$, $a$, $b$ features. 3) Extensive experiments on the benchmark dataset demonstrate that our DucoNet outperforms the state-of-the-art approaches by a large margin.

\section{Related Work}

\subsection{Image Harmonization}
As a subtask in image composition~\cite{niu2021making}, image harmonization aims to create a harmonious composite image by ensuring that the appearances of foreground and background are consistent. 
In the early stage, traditional image harmonization methods~\cite{lalonde2007using} focused on adjusting the low-level illumination and color statistics of foreground to match the background.

In recent years, deep learning based harmonization methods have brought significant advance to this research field. 
Unsupervised image harmonization methods~\cite{zhu2015learning} were initially explored using adversarial learning.
With the introduction of the first large-scale image harmonization dataset iHarmony4~\cite{dovenet}, supervised image harmonization methods~\cite{charmnet, Scs-co, S2CRNet, ssh, xing2022composite,peng2022frih,bao2022deep,zhu2022image,chen2023dense,chen2022hierarchical, PHDNet, ren2022semantic} have received increasing attention. 
Among them, some works~\cite{ssam, issam, feature_mod} designed attention modules to extract background features and adjust the foreground features through channel-wise adjustment~\cite{ssam}, semantic representation~\cite{issam, dih}, and modulation-demodulation~\cite{feature_mod}.
Additionally, some works~\cite{dovenet, rainnet, bargainnet} formulated image harmonization as domain/style translation, and employed adversarial learning~\cite{dovenet}, region-aware AdaIn~\cite{rainnet}, and contrastive loss~\cite{bargainnet} to transfer the foreground into the background domain/style. 
Moreover, some works~\cite{intrinsic, IHT, HT} introduced Retinex~\cite{land1971lightness} theory to image harmonization by decomposing the harmonization task into reflectance maintenance and illumination adjustment. 
Recently, some works~\cite{harmonizer, CDTNet} treated image harmonization as color-to-color transformation~\cite{CDTNet} or image-level regression~\cite{harmonizer}, striking a good balance between effectiveness and efficiency in high-resolution image harmonization.

Existing methods mainly rely on the correlated $RGB$ space to extract the background features and adjust the foreground features. However, the entangled $RGB$ features may increase the workload of harmonization network and impede the harmonization performance.
Our work focuses on dual color spaces (\emph{i.e.}, $RGB$ and $Lab$), by using the decorrelated $Lab$ color space to generate \textit{L}, \textit{a}, and \textit{b} control codes for feature manipulation in harmonization backbone.

\subsection{Color Spaces}
There are multiple color spaces to represent images, such as $RGB$, $Lab$, \textit{XYZ}, which can be divided into correlated and decorrelated color spaces based on whether each color channel correlates with each other. 
The correlated color space can be directly shown in different monitors and reflect the basic physics rules, for example, $RGB$ represents three primary colors of light.
However, the correlations among different color channels may prevent the critical factors to be encoded independently and complicate the color transformation~\cite{reinhard2001color}.
On the contrary, the decorrelated color space usually disentangles some critical factors (\emph{i.e.}, lightness), which may help extract the corresponding features independently.
Most works in computer vision field predominantly use $RGB$ color space. 
Nevertheless, some works also utilize multiple color spaces~\cite{peng2023u, li2021underwater} to achieve the desired effect. 

For example, in underwater image enhancement~\cite{peng2023u, li2021underwater, ma2022wavelet, zhang2022underwater}, it is important to incorporate multiple color spaces to enhance model capabilities.
Among them, Peng~\emph{et al.}\cite{peng2023u} integrated $RGB$, $Lab$, and $LCH$ color spaces into a loss function to improve the contrast and saturation of the enhanced image. 
Li~\emph{et al.}~\cite{li2021underwater} proposed a multi-color encoder to enrich the diversity of feature representations by incorporating the characteristics of $RGB$, $HSV$, and $Lab$ color spaces into a unified structure. 
Zhang~\emph{et al.}~\cite{zhang2022underwater} studied the near-independent properties of $Lab$ color space, and proposed an adaptive method to enhance the contrast and saturation in $RGB$ color.
In grayscale image coloring, Wan~\emph{et al.}~\cite{wan2020automated} utilized the $RGB$ color space to colorize the initialized super-pixel, and then employed the \textit{YUV} color space for color propagation to achieve a balance between efficiency and effectiveness. 
In video tracking, Lai~\emph{et al.}~\cite{lai2020mast} investigated loss designation in terms of different color spaces (\emph{e.g.}, $RGB$, $Lab$, and \textit{HSV}), revealing that the decorrelated color space could force models to learn more robust features.

Our work is the first deep image harmonization method using multiple color spaces. Specifically, we extract disentangled $L$, $a$, $b$ features from decorrelated $Lab$ color space, to supplement the entangled $RGB$ features extracted from correlated $RGB$ color space. 

\subsection{Dynamic Neural Network}
Dynamic neural networks aim to dynamically adjust the model parameters or structures to cope with different conditions, which can improve the generalization and representation ability of models.

For dynamic neural networks with dynamic parameters, Chen~\emph{et al.}~\cite{dynamic_conv} were the first to propose dynamic convolution, which aggregates multiple convolution kernels based on attention weight. 
CondConv~\cite{condconv} introduced the idea of learning sample-dependent convolution kernels to replace original convolution layers, resulting in improved model performance for classification and detection tasks. 
PAC~\cite{PAC} proposed the pixel-adaptive convolution operation by combining learnable local pixel features with the filter weights to change the standard convolution operation. 
In terms of dynamic neural networks with dynamic structures, MSDNet~\cite{huang2017multi} proposed a multi-scale DenseNet with an early-exit strategy that decides when to exit the network for different samples. 
ATC~\cite{graves2016adaptive} developed an algorithm that enables recurrent neural networks to learn the number of computational steps between receiving an input and emitting an output, making previously inaccessible problems manageable.

In our $Lab$ control module, inspired by StyleGANv2~\cite{styleganv2}, we use \textit{L}, \textit{a}, and \textit{b} control codes to generate dynamic convolution kernels for feature manipulation, which falls within the scope of dynamic parameters. 
This approach enables us to adjust the decoder features in the harmonization backbone using the \textit{L}, \textit{a}, and \textit{b} control codes.

\section{Method} \label{section:Method}

In this section, we will set forth to our DucoNet. 
In detail, we will first briefly introduce our overall framework in~\Cref{Overview}, and our used harmonization backbone in~\Cref{Backbone Network}.
In~\Cref{Color Style Encoder}, we will detail the process to extract the \textit{L}, \textit{a}, \textit{b} control codes.
In~\Cref{Color transfer Module}, we will describe how our $Lab$ control module ($Lab$-CM) exploits the \textit{L}, \textit{a}, \textit{b} control codes to adjust the decoder features in the harmonization backbone.

\begin{figure*}[ht]
  \centering
  \includegraphics[width=0.80\linewidth]{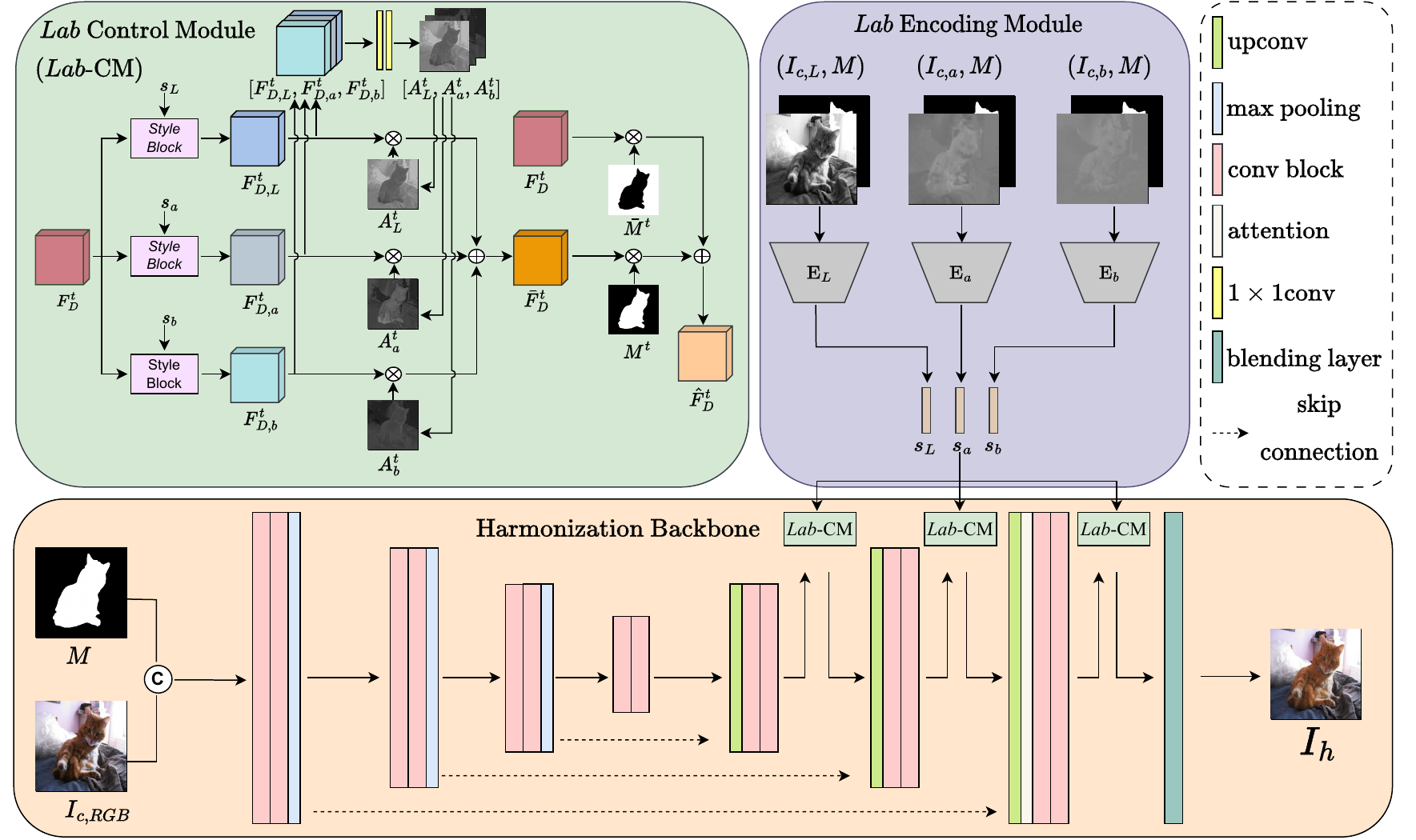}
  \caption{The illustration of our harmonization network with Dual Color Spaces (DucoNet). 
  Given a composite image $\bm{I}_{c}$ and its foreground mask $\bm{M}$, the harmonization backbone~\cite{issam} takes \textit{RGB} channels of composite image~($\bm{I}_{c,RGB}$) concatenated with $\bm{M}$ as input, and generates the harmonized image $\bm{I}_{h}$. 
  In \textit{Lab} encoding module, three encoders extract control codes $\bm{s}_{\textit{L}}$, $\bm{s}_{\textit{a}}$, and $\bm{s}_{\textit{b}}$ from \textit{L}, \textit{a}, and \textit{b} channels of composite image~$\bm{I}_{c, \textit{L}}, \bm{I}_{c, \textit{a}}, \bm{I}_{c, \textit{b}}$, respectively,
  which are used to manipulate the decoder feature maps in the harmonization backbone. 
  We insert \textit{Lab} control module (\textit{Lab}-CM) into each decoder layer. For the $t$-th decode feature map $\bm{F}_{D}^{t}$ output from the $t$-th decoder layer, 
  we use $\bm{s}_{\textit{L}}$, $\bm{s}_{\textit{a}}$, and $\bm{s}_{\textit{b}}$ to manipulate $\bm{F}_{D}^{t}$ independently through style blocks~\cite{styleganv2}. 
  Then, three manipulated decoder feature maps are fused as $\bm{\bar{F}}_{D}^{t}$ with learnt pixel-wise weights. 
  Finally, the foreground of $\bm{\bar{F}}_{D}^{t}$ and the background of $\bm{{F}}_{D}^{t}$ are combined as $\bm{\hat{F}}_{D}^{t}$ and sent back to the decoder to produce the harmonized image $\bm{I}_{h}$.}
  \Description{}
  \label{fig:framework}
  %\vspace{-5pt}
\end{figure*}

\subsection{Overview}\label{Overview}

Given a composite image $\bm{I}_{c}$ and its foreground mask $\bm{M}$, the goal of image harmonization is adjusting the foreground of $\bm{I}_{c}$ and producing the harmonized image $\bm{I}_{h}$ as output. 
Prior works~\cite{issam, CDTNet, dovenet, harmonizer, IHT} only use the composite image in the $RGB$ color space as input. 
However, $RGB$ color space is a correlated color space, which may increase the workload of existing methods to disentangle independent factors (\emph{e.g.}, lightness, orthogonal colors), potentially complicating the harmonization process.
Considering that the decorrelated $Lab$ color space contains disentangled color and illumination statistics, we additionally use the composite image with $Lab$ channels as input to help improve the  harmonization performance. 

As shown in \Cref{fig:framework}, the overall framework consists of three parts: the harmonization backbone, the $Lab$ encoding module, and the $Lab$ control module. 
Following previous works~\cite{issam, ssam}, the harmonization backbone uses the composite image with $RGB$ channels~$\bm{I}_{c, RGB} \in \mathbb{R}^{H \times W \times 3}$ concatenated with the foreground mask~$\bm{M} \in \mathbb{R}^{H \times W \times 1}$ as input. We have also tried using $Lab$ color space in harmonization backbone, but the results are compromised (see \Cref{section:Ablation Study}). Therefore, we still use $RGB$ color space in harmonization backbone.
Considering the effectiveness and efficiency, we adopt iSSAM~\cite{issam} as our harmonization backbone, which can also be easily replaced by other harmonization backbones.
For the $Lab$ encoding module, we use the composite image with $Lab$ channels $\bm{I}_{c, \textit{Lab}} \in \mathbb{R}^{H \times W \times 3}$ concatenated with the foreground mask~$\bm{M}$ as input. 
Considering that the $L$, $a$, and $b$ channels are near-independent, we process different channels $\bm{I}_{c, \textit{L}}$, $\bm{I}_{c, \textit{a}}$, $\bm{I}_{c, \textit{b}} \in \mathbb{R}^{H \times W \times 1}$ using three encoders $E_L$, $E_a$, $E_b$ separately to obtain the corresponding $L$, $a$, and $b$ control codes $\bm{s}_{\textit{L}}, \bm{s}_{\textit{a}}, \bm{s}_{\textit{b}} \in \mathbb{R}^{d_{s}},d_{s}=256$.
$Lab$ control module uses $L$, $a$, and $b$ control codes to adjust the decoder feature maps in the harmonization backbone. 
Finally, the decoder of harmonization backbone outputs the harmonized image $\bm{I}_{h}$, which is supervised by the ground-truth image $\bm{I}_{g}$ using $L_1$ loss $\mathcal{L} = || \bm{I}_{h} - \bm{I}_{g} ||_1$. 

\subsection{Harmonization Backbone}\label{Backbone Network}

The choice of harmonization backbones should balance effectiveness and efficiency simultaneously.
Therefore, we opt for iSSAM~\cite{issam} as our harmonization backbone, which is framed as a U-Net~\cite{ronneberger2015u} with four encoder layers and three decoder layers.
The first three encoder layers output features, which are connected with the corresponding decoder layers via skip connections to preserve the encoded information.
To tailor for image harmonization, an Spatial-Separated Attention Module~\cite{ssam} and a blending layer~\cite{issam} are inserted to the last decoder layer. For more details, please refer to iSSAM~\cite{issam}.  

As mentioned earlier, the harmonization backbone still uses the composite image with $RGB$ channels~$\bm{I}_{c, \textit{RGB}} \in \mathbb{R}^{H \times W \times 3}$ concatenated with the foreground mask~$\bm{M}$ as input, and outputs the harmonized result $\bm{I}_{h}$.
To adjust the decoder feature maps with the $L$, $a$, and $b$ control codes, each decoder feature map is sent into our $Lab$-CM along with the $L$, $a$, and $b$ control codes, which allows disentangled $L$, $a$, $b$ features to help produce more harmonious images. 
The details of $Lab$-CM will be introduced in \Cref{Color transfer Module}.

\subsection{\textit{Lab} Encoding Module}\label{Color Style Encoder}

The $RGB$ color space has been well explored in image harmonization tasks~\cite{dovenet, ssam, issam, intrinsic, IHT, rainnet, feature_mod, bargainnet, harmonizer, CDTNet}. Due to the correlation among $RGB$ channels, the extracted RGB features may not disentangle the independent factors (\emph{e.g.}, lightness, orthogonal colors) effectively. 
Thus, we additionally use the decorrelated $Lab$ color space to supplement $RGB$ color space. 
As introduced in \Cref{section:Introduction}, $L$, $a$, and $b$ channels in $Lab$ color space represent lightness, the spectrum from green to red, and the spectrum from blue to yellow, respectively.

In the $Lab$ color space, we attempt to obtain the control code of each channel using the respective control encoder. 
Each encoder $\text{E}_{\textit{L}}$, $\text{E}_{\textit{a}}$, and $\text{E}_{\textit{b}}$) in the $Lab$ encoding module has the same structure as the encoder of the harmonization backbone, followed by a pooling layer and a fully-connected layer. Each encoder extracts the independent feature from one channel, which serves as the control code to manipulate the decoder feature maps through our $Lab$ control module ($Lab$-CM).
In detail, we first convert the composite image from $RGB$ color space $\bm{I}_{c, RGB}\in\mathbb{R}^{H\times W\times 3}$ to $Lab$ color space $\bm{I}_{c, Lab}\in\mathbb{R}^{H\times W\times 3}$, and obtain three separate channels $\bm{I}_{c, \textit{L}}$, $\bm{I}_{c, \textit{a}}$, and $\bm{I}_{c, \textit{b}} \in\mathbb{R}^{H\times W\times 1}$. 
These three single-channel composite images are concatenated with the $\bm{M}$ and delivered to the corresponding control encoders to yield the corresponding control code.

By taking the $L$ channel~$\bm{I}_{c, \textit{L}}$ as an example, the \textit{L} control code $\bm{s}_{\textit{L}}$ is generated through the following steps. 
We first scale the range of $\bm{I}_{c, \textit{L}}$ to $[0, 1]$, and then concatenate it with $\bm{M}$ as input. The concatenation is sent into E$_{\textit{L}}$ to produce the feature map $\bm{F}_{L}$, which is then transformed into the $\textit{L}$ control code $\bm{s}_{L}$  through one pooling layer $\text{AvgPool}$ and one fully connected layers $\text{FC}_{\textit{L}}$.
The whole process for generating $L$, $a$, and $b$ control codes can be formulated as
\begin{equation}
    \begin{aligned}
      \bm{F}_{\textit{L}} &= \text{E}_{\textit{L}}(\bm{I}_{c, \textit{L}}, \bm{M}), \quad \bm{s}_{\textit{L}} = \text{FC}_{\textit{L}}(\text{AvgPool}(\bm{F}_{\textit{L}})), \\
      \bm{F}_{\textit{a}} &= \text{E}_{\textit{a}}(\bm{I}_{c, \textit{a}}, \bm{M}), \quad \bm{s}_{\textit{a}} = \text{FC}_{\textit{a}}(\text{AvgPool}(\bm{F}_{\textit{a}})), \\
      \bm{F}_{\textit{b}} &= \text{E}_{\textit{b}}(\bm{I}_{c, \textit{b}}, \bm{M}), \quad \bm{s}_{\textit{b}} = \text{FC}_{\textit{b}}(\text{AvgPool}(\bm{F}_{\textit{b}})).
    \end{aligned}
\end{equation}

With three control encoders, we get three control codes $\bm{s}_{\textit{L}}, \bm{s}_{\textit{a}}$, and $\bm{s}_{\textit{b}}$ corresponding to three channels. 
They encode the independent factors of color and illumination statistics from the composite image in $Lab$ color space, which can further provide guidance for decoder feature manipulation in our $Lab$ control module.

\subsection{\textit{Lab} Control Module}\label{Color transfer Module}
Our $Lab$ Control Module ($Lab$-CM) aims to migrate useful information from the decorrelated $Lab$ color space to the $RGB$ color space, by using three control codes to manipulate the decoder feature maps in the harmonization backbone. 
Recall that our harmonization backbone has three decoder layers and the output feature map from the $t$-th decoder layer is denoted as $\bm{F}_{D}^{t}$. We insert $Lab$-CM after each decoder layer. 
For the $t$-th decoder layer, $Lab$-CM takes $\bm{F}_{D}^{t}$ along with $L$, $a$, $b$ control codes as input, producing the $Lab$-enhanced decoder feature map~$\bm{\hat{F}}_{D}^t$. Precisely, we first use three control codes to get three manipulated feature maps independently, and then fuse them using learnt pixel weights.  

\noindent\textbf{Feature Map Manipulation: }By taking the decoder feature map $\bm{F}_{D}^{1}$ from the first decoder layer as an example, we attempt to use three control codes $\bm{s}_{\textit{L}}, \bm{s}_{\textit{a}}$, and $\bm{s}_{\textit{b}}$ to manipulate $\bm{F}_{D}^{1}$ independently and obtain three manipulated decoder feature maps. 
In this work, we adopt the style block proposed in StyleGANv2~\cite{styleganv2}, which is essentially dynamic convolution. The style block produces dynamic convolution kernel using the control code and apply it to the decoder feature map. 

Specifically, for each color channel $c$ from \{\textit{L, a, b}\}, we have one $3\times 3$ base convolution kernel $\bm{W}_{c}$, and use control code $\bm{s}_{\textit{c}}$ to dynamically scale the input channels of $\bm{W}_{c}$. We first project $\bm{s}_{\textit{c}}$ to a scale vector $\bm{u}_{\textit{c}}$ using two fully-connected layers, in which $\bm{u}_{\textit{c}}$ contains the scales for each input channel. The scaling process is represented by
\begin{eqnarray}
      \hat{w}_{c}^{i, j, k} =  u_{c}^{i} \cdot {w}_{c}^{i, j, k},
\end{eqnarray}
in which $w_{c}^{i, j, k}$ is the $(i,j,k)$-th entry in $\bm{W}_{c}$ with $i, j, k$ enumerating the input channel, output channel, and the spatial location respectively. $u_{c}^{i}$ is the $i$-th entry in $\bm{u}_{\textit{c}}$, representing the scale for the $i$-th input channel. Then, we normalize $\hat{w}_{c}^{i, j, k}$ as
\begin{eqnarray}
      \bar{w}_{c}^{i, j, k} = \hat{w}_{c}^{i, j, k} \bigg/ \sqrt{(\sum_{i, k} \hat{w}_{c}^{i, j, k})^2 + \epsilon},
\end{eqnarray}
where $\epsilon$ is a small constant to prevent numerical errors. $\bar{w}_{c}^{i, j, k}$ form the dynamic convolution kernel $\bm{\bar{W}}_{c}$, which acts upon the decoder feature map $\bm{F}_{D}^{1}$ to produce the manipulated feature map $\bm{F}_{D, \textit{L}}^{1}$. For more details of the style block, please refer to StyleGANv2~\cite{styleganv2}.

With three control codes, we can get three manipulated feature maps~$\bm{F}_{D, \textit{L}}^{1}$, $\bm{F}_{D, \textit{a}}^{1}$, $\bm{F}_{D, \textit{b}}^{1}$. By using  P$_{c}$ to denote the style block for the color channel $c$, the feature map manipulation can be formulated as
\begin{equation}
    \begin{aligned}
      \bm{F}_{D, \textit{L}}^{1} \!=\! \text{P}_\textit{L}(\bm{F}_{D}^{1}, \bm{s}_{\textit{L}}), \quad
      \bm{F}_{D, \textit{a}}^{1} \!=\! \text{P}_\textit{a}(\bm{F}_{D}^{1}, \bm{s}_{\textit{a}}), \quad
      \bm{F}_{D, \textit{b}}^{1} \!=\! \text{P}_\textit{b}(\bm{F}_{D}^{1}, \bm{s}_{\textit{b}}).
    \end{aligned}
\end{equation}

\noindent\textbf{Feature Map Fusion: } Considering that $L$, $a$, $b$ channels may contribute differently to various images or even various pixels, we learn pixel-wise weights $\{\bm{A}_{\textit{L}}^{1}, \bm{A}_{\textit{a}}^{1}, \bm{A}_{\textit{b}}^{1}\}$ for three channels when fusing three manipulated feature maps $\{\bm{F}_{D, \textit{L}}^{1}, \bm{F}_{D, \textit{a}}^{1}, \bm{F}_{D, \textit{b}}^{1}\}$. Specifically, we concatenate three manipulated feature maps and send them to $G^1$:
\begin{equation}
    [\bm{A}_{\textit{L}}^{1}, \bm{A}_{\textit{a}}^{1}, \bm{A}_{\textit{b}}^{1}] = \text{G}^{1}\left([\bm{F}_{D, \textit{L}}^{1}, \bm{F}_{D, \textit{a}}^{1}, \bm{F}_{D, \textit{b}}^{1}]\right),
\end{equation}
where G$^{1}$ is constructed by a $1 \times 1$ convolution layer and a softmax layer, $\{\bm{A}_{\textit{L}}^{1}, \bm{A}_{\textit{a}}^{1}, \bm{A}_{\textit{b}}^{1}\}$ are single-channel weight maps. %, and $\bm{W}_{f, p}^{1}$, $p$ $\in$ \{\textit{L, a, b}\} shares the same resolution as $\bm{F}_{D, p}^{1}$, $p$ $\in$ \{\textit{L, a, b}\}. 
After that, we fuse three manipulated feature maps~($\bm{F}_{D, \textit{L}}^{1}$, $\bm{F}_{D, \textit{a}}^{1}$, $\bm{F}_{D, \textit{b}}^{1}$) using the predicted pixel-wise weights. Note that we only manipulate the foreground feature map, aiming to make it compatible with the background. Thus, the original background feature map in $\bm{F}_{D}^{1}$ is preserved. The above process is represented by
\begin{equation}\label{equation_4}
    \begin{aligned}
    \bm{\bar{F}}_{D}^{1} &= \bm{F}_{D, \textit{L}}^{1} \circ \bm{A}_{\textit{L}}^{1} + \bm{F}_{D, \textit{a}}^{1} \circ \bm{A}_{\textit{a}}^{1} + \bm{F}_{D, \textit{b}}^{1} \circ \bm{A}_{\textit{b}}^{1}, \\
    \bm{\hat{F}}_{D}^{1} &= \bm{\bar{F}}_{D}^{1} \circ \bm{M}^{1} + (1-\bm{M}^{1}) \circ \bm{F}_{D}^{1}, 
    \end{aligned}
\end{equation}
where $\circ$ means element-wise product and $\bm{\hat{F}}_{D}^{1}$ is the final $Lab$-enhanced feature map. 

Similar steps can be applied to decoder feature maps $\bm{F}_{D}^{2}$ and $\bm{F}_{D}^{3}$ to get the $Lab$-enhanced feature maps $\bm{\hat{F}}_{D}^{2}$ and $\bm{\hat{F}}_{D}^{3}$. The $Lab$-enhanced feature maps are sent back to the decoder of the harmonization backbone to generate the final harmonized image $\bm{I}_{h}$.

\begin{figure*}[htbp]
  \centering
  \includegraphics[width=0.94\linewidth]{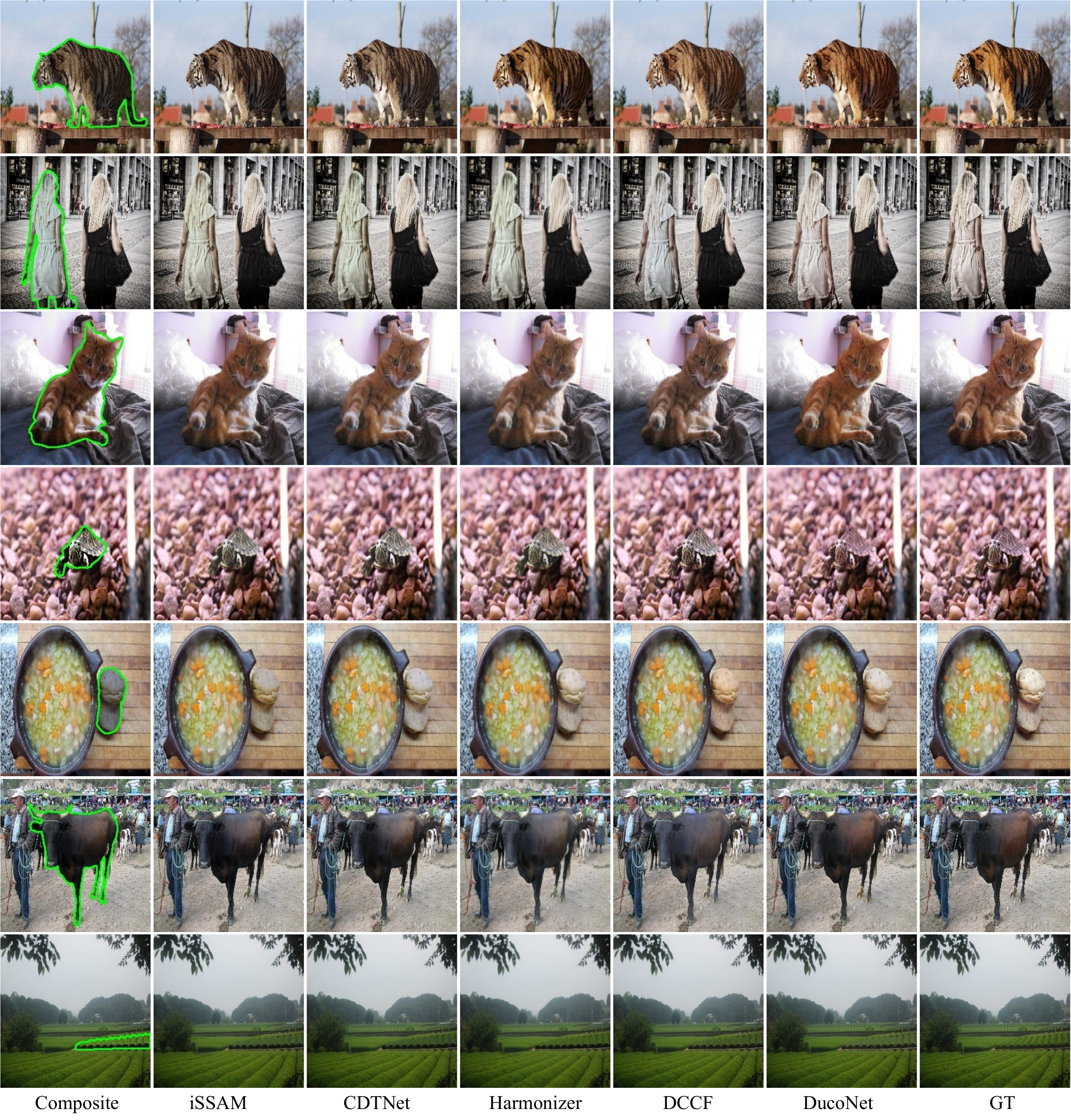}
  \caption{From left to right, we show the composite image ( foreground outlined  in green), the harmonized results of iSSAM~\cite{issam}, CDTNet~\cite{CDTNet}, Harmonizer ~\cite{harmonizer}, DCCF~\cite{DCCF}, our DucoNet, and the ground-truth in iHarmony4~\cite{dovenet} dataset. Best viewed in color and zoom in.}
  \Description{}
  \label{fig:baseline_IHD}
\end{figure*}

\begin{table*}[t]
    \centering
    \setlength\tabcolsep{0.9pt}
    \resizebox{\textwidth}{!}{
    \begin{tabular}{c|ccc|ccc|ccc|ccc|ccc}
    \toprule
        \multirow{2}{4em}{\textbf{Method}} & \multicolumn{3}{c|}{\textbf{All}} & \multicolumn{3}{c|}{\textbf{HCOCO}} & \multicolumn{3}{c|}{\textbf{HFlickr}} & \multicolumn{3}{c|}{\textbf{HAdobe5k}} & \multicolumn{3}{c}{\textbf{Hday2night}}  \\
        \cline{2-16}
        &\textbf{MSE $\downarrow$} & \textbf{fMSE $\downarrow$ } & \textbf{PSNR $\uparrow$ } 
        &\textbf{MSE $\downarrow$} & \textbf{fMSE $\downarrow$ } & \textbf{PSNR $\uparrow$ }
        &\textbf{MSE $\downarrow$} & \textbf{fMSE $\downarrow$ } & \textbf{PSNR $\uparrow$ }
        &\textbf{MSE $\downarrow$} & \textbf{fMSE $\downarrow$ } & \textbf{PSNR $\uparrow$ }
        &\textbf{MSE $\downarrow$} & \textbf{fMSE $\downarrow$ } & \textbf{PSNR $\uparrow$ } \\ 
    \midrule
        Composite images & 172.47 & 1387.30 & 31.63 & 69.37 & 1013.27 & 33.94 & 264.35 & 1612.59 & 28.32 & 345.54 & 2137.07 & 28.16 & 109.65 & 1443.05 & 34.01 \\
        DoveNet~\cite{dovenet} & 52.36  & 549.96  & 34.75  & 36.72  & 554.55  & 35.83  & 133.14  & 823.64  & 30.21  & 52.32  & 383.91  & 34.34  & 54.05  & 1075.42  & 35.18   \\ 
        RainNet~\cite{rainnet}& 40.29  & 469.60  & 36.12  & 31.12  & 535.40  & 37.08  & 117.59  & 751.12  & 31.64  & 42.85  & 320.43  & 36.22  & 47.24  & 852.12  & 34.83   \\ 
        Instrinsic~\cite{intrinsic} & 38.71  & 400.29  & 35.90  & 24.92  & 416.38  & 37.16  & 105.13  & 716.60  & 31.34  & 43.02  & 284.21  & 35.20  & 55.53  & 797.04  & 35.96   \\ 
        IHT~\cite{IHT} & 27.89  & 295.56  & 37.94  & 14.98  & 274.67  & 39.22  & 67.88  & 471.04  & 33.55  & 36.83  & 242.57  & 37.17  & 49.67  & 736.55  & 36.38   \\ 
        iSSAM~\cite{issam} & 24.64  & 262.67  & 37.95  & 16.48  & 266.14  & 39.16  & 69.68  & 443.63  & 33.56  & 22.59  & 166.19  & 37.24  & 40.59  & 591.07  & 37.72   \\ 
        CDTNet~\cite{CDTNet} & 23.75  & 252.05  & 38.23  & 16.25  & 261.29  & 39.15  & 68.61  & 423.03  & 33.55  & 20.62  & 149.88  & 38.24  & 36.72  & 549.47  & 37.95   \\ 
        Harmonizer~\cite{harmonizer} & 24.26  & 280.51  & 37.84  & 17.34  & 298.42  & 38.77  & 64.81  & 434.06  & 33.63  & 21.89  & 170.05  & 37.64  & \textbf{33.14}  & 542.07  & 37.56   \\ 
        DCCF~\cite{DCCF} & 22.05  & 266.49  & 38.50  & 14.87  & 272.09  & 39.52  & 60.41  & 411.53  & 33.94  & 19.90  & 175.82  & 38.27  & 49.32  & 655.43  & 37.88   \\ 
        \cline{1-16}
        DucoNet & \textbf{18.47} & \textbf{212.53} & \textbf{39.17} & \textbf{12.12} & \textbf{211.25} & \textbf{40.23} & \textbf{51.71} & \textbf{353.81} & \textbf{34.65} & \textbf{17.06} & \textbf{141.55} & \textbf{38.87} & 38.70 & \textbf{527.07} & \textbf{38.11} \\
    \bottomrule
    \end{tabular}}
    \caption{Comparison of different methods with image size 256 $\times$ 256 on iHarmony4. $\downarrow$ (\emph{resp.}, $\uparrow$) indicates that lower (\emph{resp.}, higher) values are better. The best results are highlighted in bold face. }
    \label{table:1}
\end{table*}

\begin{table}[t]
    \centering
    \setlength\tabcolsep{7.8pt}
    \resizebox{\columnwidth}{!}{
    \begin{tabular}{c|ccc}
    \toprule
        \textbf{Method} & \textbf{MSE $\downarrow$ } & \textbf{fMSE $\downarrow$ }  & \textbf{PSNR $\uparrow$}  \\ 
    \midrule
         Composite images & 352.05 & 2122.37 & 28.10  \\ 
         iSSAM~\cite{issam}            & 25.03 & 168.85 & 38.29  \\
         CDTNet-256(sim)~\cite{CDTNet} & 31.15 & 195.93 & 37.65  \\
         CDTNet-256~\cite{CDTNet}      & 21.24 & 152.13 & 38.77  \\
         Harmonizer~\cite{harmonizer}  & 20.12 & 150.99 & 38.45 \\
         DCCF~\cite{DCCF}              & 21.12 & 171.17 & 38.38  \\
        \cline{1-4}
         DucoNet             & \textbf{10.94} & \textbf{80.69} & \textbf{41.37} \\
    \bottomrule
    \end{tabular}}
    \caption{Comparison of different methods with image size $1024 \times 1024$ on HAdobe5k. $\downarrow$ (\emph{resp.}, $\uparrow$) indicates that lower (\emph{resp.}, higher) values are better. The best results are denoted in bold face.}
    \label{table:2}
\end{table}

\section{Experiments}
\subsection{Datasets and Evaluation Metrics}
\subsubsection{Dataset} 
Following previous image harmonization works, we conduct experiments on the benchmark dataset iHarmony4~\cite{dovenet} to evaluate the effectiveness of our DucoNet, where the iHarmony4~\cite{dovenet} has been widely used in supervised image harmonization. 
In detail, iHarmony4~\cite{dovenet} consists of four sub-datasets, including HFlickr, Hday2night, HCOCO, and HAdobe5K, with 73,146 samples in total.
For each sample in iHarmony4~\cite{dovenet}, it includes a composite image, its foreground mask, and the corresponding ground-truth image.

We perform both low-resolution and high-resolution image harmonization based on iHarmony4. 
For low-resolution harmonization, we conduct experiments with image size $256 \times 256$ following previous works~\cite{issam}.
For high-resolution harmonization, we follow the experimental setting in CDTNet~\cite{CDTNet}. Specifically, we perform training and testing based on the HAdobe5k dataset with image size $1024 \times 1024$. 
Moreover, we also evaluate our trained model on 100 high-resolution real composite images collected in CDTNet~\cite{CDTNet}. 
Since real composite images have no ground-truth image for evaluation, we present the user study results.

\subsubsection{Evaluation Metrics} 
We adopt the evaluation metrics which are commonly used in previous image harmonization works~\cite{dovenet,IHT,issam,CDTNet,harmonizer,DCCF}, including MSE (Mean-Square-Error), fMSE (foreground Mean-Square-Error), and PSNR (Peak Signal to Noise Ratio).

\subsection{Implementation Details}

Our network is implemented with PyTorch 1.10.1, optimized by Adam optimizer with initial learning rate as $1\times 10^{-3}$.
The batch size is set as 64 and we train our DucoNet for 120 epochs in total.
The learning rate decay starts at epoch 105 and epoch 115 with a decay factor of 10. 
The hardware devices used for training are Intel(R) Xeon(R) Silver 4116 CPU, with 128GB memory and two NVIDIA GeForce RTX 3090 GPUs.
More details about the implementation can be found in Supplementary.

\subsection{Comparison with Start-of-the-Art Methods}
\noindent\textbf{Low-resolution Harmonization: } We compare our method with the existing methods. 
In the low-resolution setting with image size $256 \times 256$, we compare our method with DoveNet~\cite{dovenet}, RainNet~\cite{rainnet}, Instrinsic~\cite{intrinsic}, IHT (Image Harmonization with Transformer)~\cite{IHT}, iSSAM~\cite{issam}, CDTNet~\cite{CDTNet}, Harmonizer~\cite{harmonizer}, and DCCF~\cite{DCCF}. The experiment results are copied from original papers or reproduced with the released models.

In \Cref{table:1}, we report the results on four sub test sets and the whole test set in the low-resolution setting.
For the results on the whole test set, our DucoNet outperforms the SOTA method by a large margin.
Specifically, our DucoNet achieves 15.68\% relative improvement over CDTNet~\cite{CDTNet} in terms of fMSE and 16.23\% relative improvement over DCCF~\cite{DCCF} in terms of MSE. 
Considering each sub test set, our DucoNet achieves the best results on HCOCO, HFlickr, and HAdobe5k, which indicates the generation ability our method.
On Hday2night, our method achieves the best results in terms of fMSE and PSNR, and the third best result for MSE, probably due to the small-scale training set and test set (only 311 images for training and 133 image for test). 

We further visualize the harmonized results of different methods in \Cref{fig:baseline_IHD}.
It can be seen that our method can produce more visually appealing and harmonious results, that are closer to the ground-truth real images. These visualisation results again demonstrate the effectiveness of our proposed method. 

\noindent\textbf{High-resolution Harmonization: } 
Recently, there are also a few works that focus on high-resolution image harmonization.
In the high-resolution setting with image size $1024 \times 1024$, we compare our  DucoNet with iSSAM~\cite{issam}, CDTNet~\cite{CDTNet}, Harmonizer~\cite{harmonizer}, DCCF~\cite{DCCF} in HAdobe5k subset with image size $1024 \times 1024$. 
CDTNet-256 is the CDTNet~\cite{CDTNet} model with the input size of harmonization backbone being $256\times 256$, and CDTNet-256(sim) is a simplified version of CDTNet-256. 
The experimental results for DCCF, CDTNet-256 and CDTNet-256(sim) are copied from the corresponding paper. 
Harmonizer did not report their results in the same high-resolution setting as CDTNet~\cite{CDTNet}, so we train the corresponding models on HAdobe5k training set with image size $1024 \times 1024$ for fair comparison. 

In \Cref{table:2}, we report the results on HAdobe5k in the high-resolution image harmonization setting.
Our DucoNet outperforms all the baselines by a large margin in terms of all evaluation metrics in high-resolution image harmonization.
Specifically, our DucoNet achieves 45.63\% relative improvement over Harmonizer~\cite{harmonizer} in terms of MSE and achieves 46.56\% relative improvement over Harmonizer~\cite{harmonizer} in terms of fMSE.

\subsection{Ablation Study}\label{section:Ablation Study}

As described in \Cref{section:Method}, our DucoNet consists of the harmonization backbone, the $Lab$ encoding module, and the $Lab$ control module ($Lab$-CM). 
In this section, we demonstrate the effectiveness of each component and each color space by ablating each component or comparing with alternatives.

The results of our ablation studies are presented in \Cref{table:3}. Firstly, when only using the harmonization backbone, we compare using the input composite image with $RGB$ channels (row 1) and using the input composite image with $Lab$ channels (row 2).
By comparing row 1 and row 2,  we see that $RGB$ channels outperforms $Lab$ channels, revealing that $RGB$ channels are still more suitable as the input for the U-Net structure. 
Note that although the inputs to the network are different, the loss and evaluation metrics are all calculated based on $RGB$ channels for fair comparison.
In detail, when using the input composite image with $Lab$, we first generate the harmonized image with $Lab$ channels and then convert it into $RGB$ color space for loss calculation and evaluation.

To evaluate the effectiveness of $Lab$ color space for feature manipulation, we treat the $Lab$ (\emph{resp.}, $RGB$) channels as a whole input in the encoding module and use a single control code in the control module, leading to the results in row 3 (\emph{resp.}, row 4). 
Comparing row 3 and row 4, we can find that the $Lab$ channels are more helpful for feature manipulation, because $Lab$ color space could supplement $RGB$ color space with extra useful guidance.

In row 5, we study a simple way to fuse $RGB$ and $Lab$ features. In particular, we treat the $Lab$ channels as a whole input in encoding module and send multi-scale encoder features to the decoder via skip-connection, in the same way as the backbone encoder. The obtained performance is worse than row 3,  which demonstrates the effectiveness of feature manipulation in our $Lab$-CM.

Furthermore, we conduct experiments by treating each individual $L$, $a$, $b$ channel as the input in the encoding module and use the single control code in the control module (row 6 \emph{v.s.} row 7 \emph{v.s.} row 8). 
Experimental results shows the $L$ channel is the most effective one among all three channels. 
To provide some insights for the importance of $L$ channel, we calculate the amount of change between the foreground area of composite image and the ground-truth image for each channel ($L$, $a$, and $b$), the average amount of change in three channels are 25.90, 3.88, and 6.65 respectively over the entire test set.
The average amount of change in $L$ channel is significantly higher than the other two channels, which corroborates that merely using $L$ channel could achieve compelling results (row 6). 

Finally, we conduct experiments to verify the effectiveness of the pixel-wise weighting strategy.
Comparing row 9 with row 10, we can find that simply averaging the manipulated feature maps~$\{\bm{F}_{D, \textit{L}}^{t}$, $\bm{F}_{D, \textit{a}}^{t}$, $\bm{F}_{D, \textit{b}}^{t}\}$ undermines the representation ability of $Lab$-CM, since $L$, $a$, $b$ channels contribute differently to the harmonization results.

\begin{table}[t]
    \centering
    \setlength\tabcolsep{3.7pt}
    \resizebox{\columnwidth}{!}{
    \begin{tabular}{c|c|c|c|ccc}
    \toprule
        \makebox[1em][c]{} & \textbf{$RGB$}& \textbf{$Lab$} & \textbf{Fusion} & \textbf{MSE $\downarrow$ } & \textbf{fMSE $\downarrow$ } & \textbf{PSNR $\uparrow$} \\ 
    \midrule
    1&    iSSAM &  -    & - & 24.64 & 262.67 & 37.95 \\
    2&      -   & iSSAM & -  & 28.13 & 296.59 & 37.20  \\
    3&    iSSAM & E($Lab$) & CM   & 19.30 & 222.22 & 38.93  \\
    4&    iSSAM & E(\textit{RGB}) & CM & 22.66 & 245.38 & 38.62   \\
    5&    iSSAM & E($Lab$) & SC   & 21.76 & 243.06 & 38.47 \\
    6&    iSSAM & E(\textit{L})   & CM & 21.43 & 234.70 & 38.72   \\
    7&    iSSAM & E(\textit{a})   & CM & 23.32 & 256.34 & 38.39   \\
    8&    iSSAM & E(\textit{b})   & CM & 23.46 & 255.29 & 38.36 \\
    9&    iSSAM & E(\textit{L,a,b}) & CM-avg & 20.45 & 227.71 & 38.88 \\
    10&   iSSAM & E(\textit{L,a,b}) & CM-pix & 18.47 & 212.53 & 39.17 \\
    \bottomrule
    \end{tabular}}
    \caption{The ablation study of our DucoNet. 
    ``iSSAM" indicates using the harmonization backbone~\cite{issam} in the corresponding color space.
    ``E(\textit{Lab})", ``E(\textit{RGB})", ``E(\textit{L})", ``E(\textit{a})", ``E(\textit{a})", and ``E(\textit{L,a,b})" indicate that we treat \textit{Lab} as a whole, \textit{RGB} as a whole, only \textit{L}, only \textit{a}, only \textit{b}, and  \textit{L,a,b} separately as input in the \textit{Lab} encoding module.
    ``SC" is short for skip-connection.
    ``CM" is short for \textit{Lab}-CM.
    ``CM-avg'' indicates average fusion.
    ``CM-pix'' indicates weighted fusion with pixel-wise weights.
    }
    \label{table:3}
\end{table}

\subsection{Visualization of Weight Map}

\begin{figure}[t]
  \centering
  \includegraphics[width=0.95\linewidth]{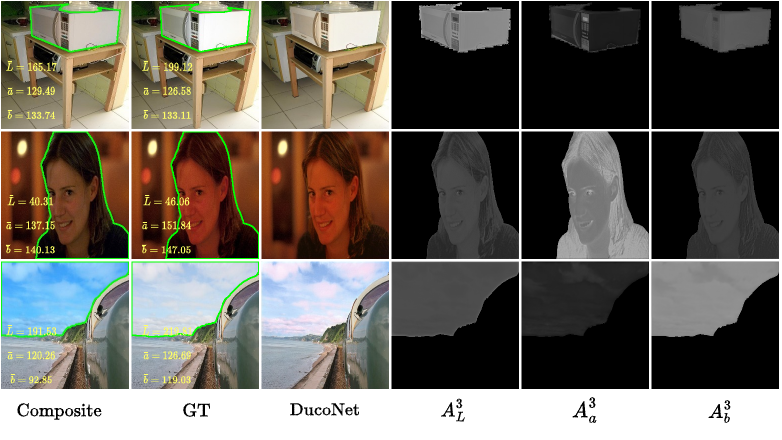}
  \caption{From left to right, we show the composite image ( foreground outlined  in green), the ground-truth, the harmonized results of our method, visualization of $\{\bm{A}_{\textit{L}}^{3}, \bm{A}_{\textit{a}}^{3}, \bm{A}_{\textit{b}}^{3}\}$ in \textit{Lab}-CM. In composite image and ground-truth image, we also show the average value in \textit{L}, \textit{a}, \textit{b} channels within the foreground region. Best viewed in color and zoom in.}
  \Description{}
  \label{fig:weight_visual}
\end{figure}

To show the effectiveness of our proposed $Lab$-CM, we visualize the weight maps $\{\bm{A}_{\textit{L}}^{3}, \bm{A}_{\textit{a}}^{3}, \bm{A}_{\textit{b}}^{3}\}$ from the third decoder layer in \Cref{fig:weight_visual}. Recall that we only manipulate the foreground region of decoder feature map and the background pixel weights do not contribute to the final output. Thus, we mask out the background pixels and only show the pixel weights in the foreground region in \Cref{fig:weight_visual}, in which brighter pixel indicates higher weight.
For composite image and ground-truth, we also show the average value in $L$, $a$, $b$ channels within the foreground region, which reflects the amount of change in each channel. 

Based on \Cref{fig:weight_visual}, we observe that the learnt weight map is closely related to the amount of change in each channel. 
Recall that $L$, $a$, and $b$ channels in $Lab$ color space represent lightness, the spectrum from green to red, and the spectrum from blue to yellow, respectively. 
When the lightness between foreground and background in the composite image is contrastively different (row 1), the value of $L$ channel would change greatly after harmonization, in which case the weight map $\bm{A}_{L}$ corresponding to the $L$ channel has the largest values. 
When the foreground object has dominant color (row 3) or the lighting has color cast (row 2),  the value of the corresponding color channel (\emph{e.g.}, red, blue) would vary greatly after harmonization, in which the corresponding weight map has the largest values (\Cref{fig:weight_visual}).

\subsection{Real Composite Images}
Following previous works, we also evaluate different methods on 100 real composite images in CDTNet~\cite{CDTNet}.  The visualization results of different baseline methods are provided in the Supplementary.
Since these real composite images do not have ground-truth image, we conduct user study to compare different methods, which is also left to the Supplementary. 

\section{Conclusion}
In this paper, we have explored image harmonization in dual color spaces, where we additionally use the decorrelated color space $Lab$ to relieve the burden of the harmonization process when compared with using $RGB$ color space alone.
We have proposed a novel network DucoNet, which manipulates the foreground of the decoder feature maps from the harmonization backbone using the control codes from $Lab$ color space. Experiments conducted on the benchmark dataset have shown that our approach significantly outperforms the state-of-the-art methods.

\begin{acks}
 The work was supported by the Shanghai Municipal Science and Technology Major / Key Project, China (Grant No. 20511100300 / 2021SHZDZX0102) and the National Natural Science Foundation of China (Grant No. 62076162).
\end{acks}

\bibliographystyle{ACM-Reference-Format}
\bibliography{acmart}

\end{document}

\title{Supplementary for Deep Image Harmonization in Dual Color Spaces}

\author{Linfeng Tan}
\orcid{0009-0003-3376-0989}
\affiliation{%
  \institution{MoE Key Lab of Artificial Intelligence,}
  \institution{Shanghai Jiao Tong University}
  \country{China}
}
\email{tanlinfeng@sjtu.edu.cn}

\author{Jiangtong Li}
\orcid{0000-0003-3873-4053}
\affiliation{%
  \institution{MoE Key Lab of Artificial Intelligence,}
  \institution{Shanghai Jiao Tong University}
  \country{China}
}
\email{keep_moving-Lee@sjtu.edu.cn}

\author{Li Niu}
\orcid{0000-0003-1970-8634}
\authornote{Corresponding author.}
\affiliation{%
   \institution{MoE Key Lab of Artificial Intelligence,}
  \institution{Shanghai Jiao Tong University}
  \country{China}
}
\email{ustcnewly@sjtu.edu.cn}

\author{Liqing Zhang}
\authornotemark[1]
\orcid{0000-0001-7597-8503}
\affiliation{%
   \institution{MoE Key Lab of Artificial Intelligence,}
  \institution{Shanghai Jiao Tong University}
  \country{China}
}
\email{zhang-lq@cs.sjtu.edu.cn}

\maketitle

\appendix

In this document, we will provide additional materials to support our main submission.
In \Cref{sec:More Details of Our method}, we will present more implementation details about our DucoNet. 
In \Cref{sec:More Visual Comparison with Baselines}, we will provide more visual comparison with baselines on both low-resolution and  high-resolution image harmonization. 
In \Cref{sec:Real Composite Images}, we will show the visualisation results on real composite images and the user study results. 
In \Cref{sec:Visualization of Abalation Studies}, we will provide the visualization results of ablation studies. 
In \Cref{sec:Failure Cases}, we will discuss the failure cases of our DucoNet.

\section{Implementation Details} \label{sec:More Details of Our method}

As mentioned in the main submission, our DucoNet consists of the harmonization backbone, the $Lab$ encoding module, and the $Lab$ control module~($Lab$-CM).
For harmonization backbone, we exploit iSSAM~\cite{issam} as our harmonization backbone due to the balance of effectiveness and efficiency. Please refer to iSSAM~\cite{issam} for more details.
For $Lab$ encoding module, $\text{E}_{\textit{L}}$, $\text{E}_{\textit{a}}$, and $\text{E}_{\textit{b}}$ have the same structure as the encoder of the harmonization backbone, where the number of the output channels are all set as 256.
The input channels and the output channels for $\text{FC}_{\textit{L}}$,$\text{FC}_{\textit{a}}$ and $\text{FC}_{\textit{b}}$ are both 256.
For $Lab$ control module~($Lab$-CM), the number of the output channels for the style blocks are set as 128, 64, and 32, respectively, which is the same as the corresponding decoder feature map.

\section{More Visual Comparison}\label{sec:More Visual Comparison with Baselines}

\begin{figure*}[htbp]
  \centering
  \includegraphics[width=0.98\linewidth]{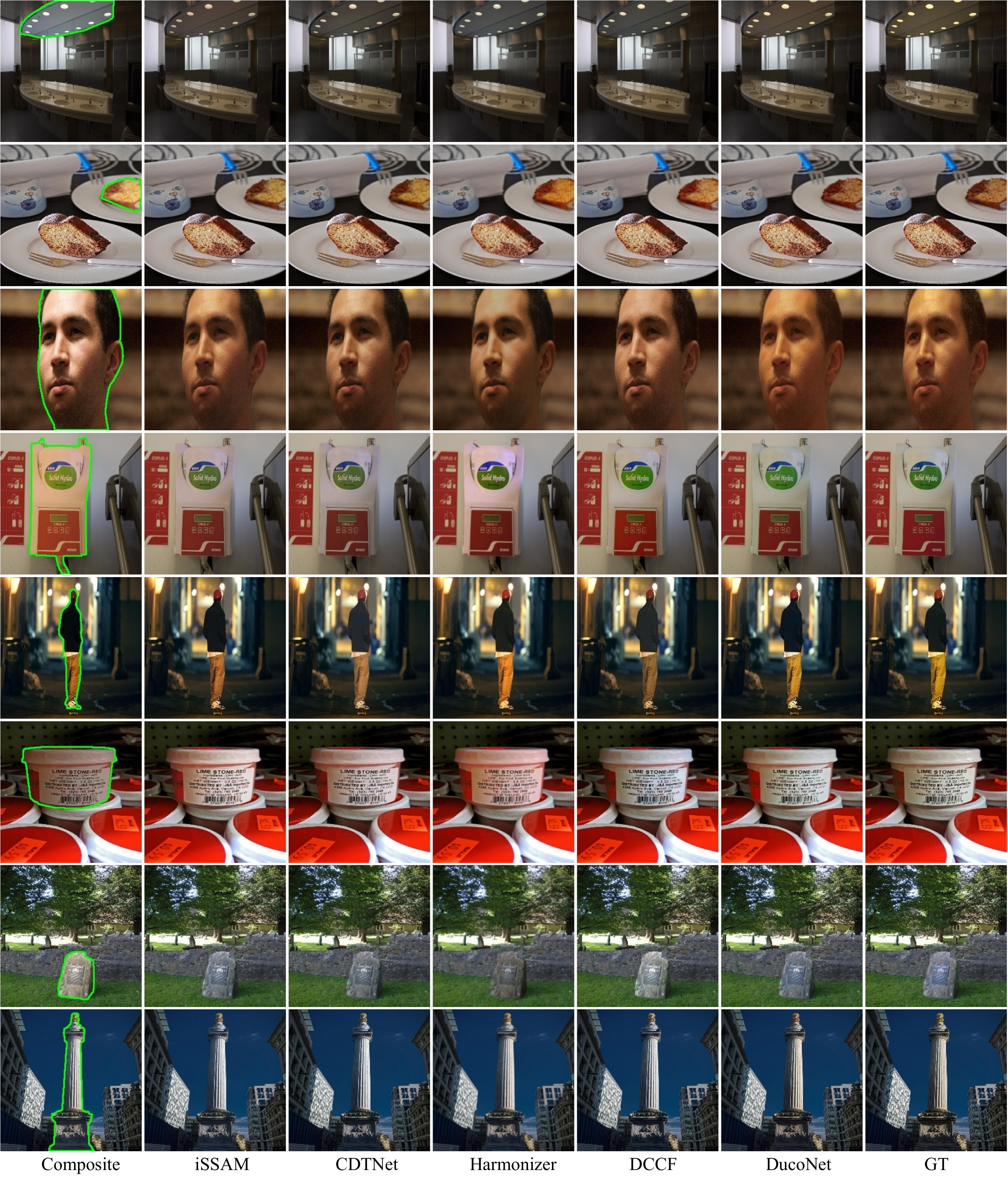}
  \caption{From left to right, we show the low-resolution composite image (foreground outlined  in green), the harmonized results of iSSAM~\cite{issam}, CDTNet~\cite{CDTNet}, Harmonizer ~\cite{Harmonizer}, DCCF~\cite{DCCF}, our DucoNet, and the ground-truth on iHarmony4~\cite{dovenet} dataset. Best viewed in color and zoom in.}
  \Description{}
  \label{fig:visual_256}
\end{figure*}

\begin{figure*}[htbp]
  \centering
  \includegraphics[width=0.98\linewidth]{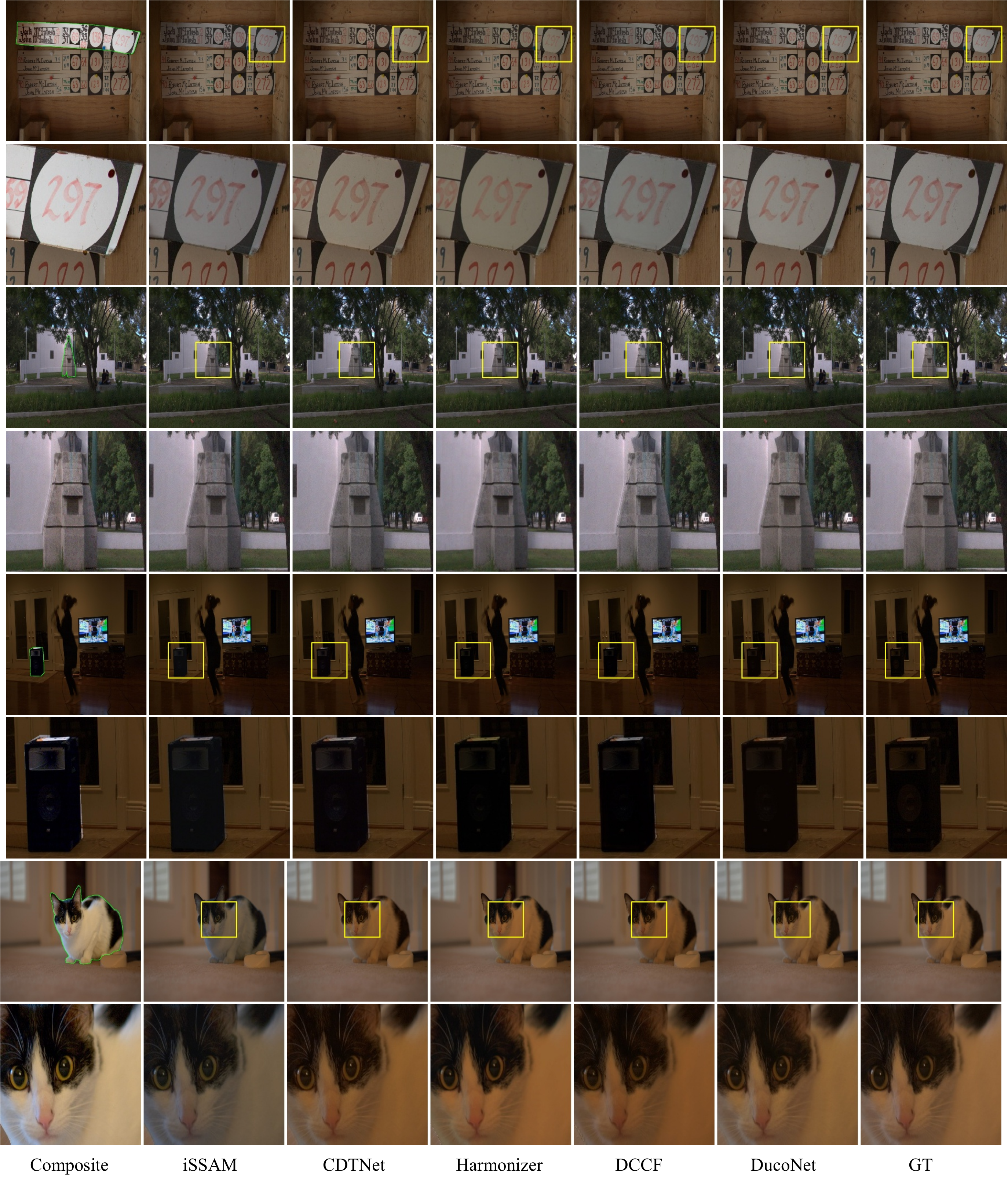}
  \caption{From left to right, we show the high-resolution composite image (foreground outlined  in green), the harmonized results of iSSAM~\cite{issam}, CDTNet~\cite{CDTNet}, Harmonizer ~\cite{Harmonizer}, DCCF~\cite{DCCF}, our DucoNet, and the ground-truth on HAdobe5k ($1024\times 1024$). The yellow boxes zoom in specific areas for better viewing.}
  \Description{}
  \label{fig:visual_1024}
\end{figure*}

To qualitatively compare with the baseline methods, including iSSAM~\cite{issam}, CDTNet~\cite{CDTNet}, Harmonizer~\cite{Harmonizer}, DCCF~\cite{DCCF}, we show more harmonization results of low-resolution image harmonization ($256 \times 256$) in \Cref{fig:visual_256} and high-resolution image harmonization ($1024 \times 1024$) in \Cref{fig:visual_1024}. 

Compared with other baseline methods, our DucoNet achieves better results in terms of both luminance and color for both low-resolution and high-resolution image harmonization. 
Our harmonization results are closer to the ground truth and more visually pleasing. These visualization results again demonstrate the effectiveness of our proposed method. 

\section{Real Composite Images}\label{sec:Real Composite Images}

For performance evaluation on real composite images, we use the 100 real composite images collected by CDTNet~\cite{CDTNet}. 
Since the real composite images have no ground-truth harmonized images, we conduct user study to compare different methods. 
Following the evaluation process in~\cite{CDTNet}, we perform comparison among the composite images and the harmonization results of five harmonization methods including iSSAM~\cite{issam}, CDTNet~\cite{CDTNet}, Harmonizer~\cite{Harmonizer}, DCCF~\cite{DCCF}, and our DucoNet.
Specifically, for each composite image, we can obtain one composite image and five harmonized results, leading to 15  pairs.
So all 100 composite images result in 1,500 image pairs.
Given each pair, 50 users are invited to identify the more harmonious one. Finally, 75,000 comparison results are collected, followed by using the Bradley-Terry (B-T) model \cite{bradley1952rank,lai2016comparative} to calculate a global ranking of all methods.
As presented in \Cref{table:hr100_user_study}, our DucoNet achieves the highest B-T score, since our DucoNet takes advantage of both decorrelated and correlated color spaces to obtain better results.

\begin{figure*}[htbp]
  \centering
  \includegraphics[width=0.98\linewidth]{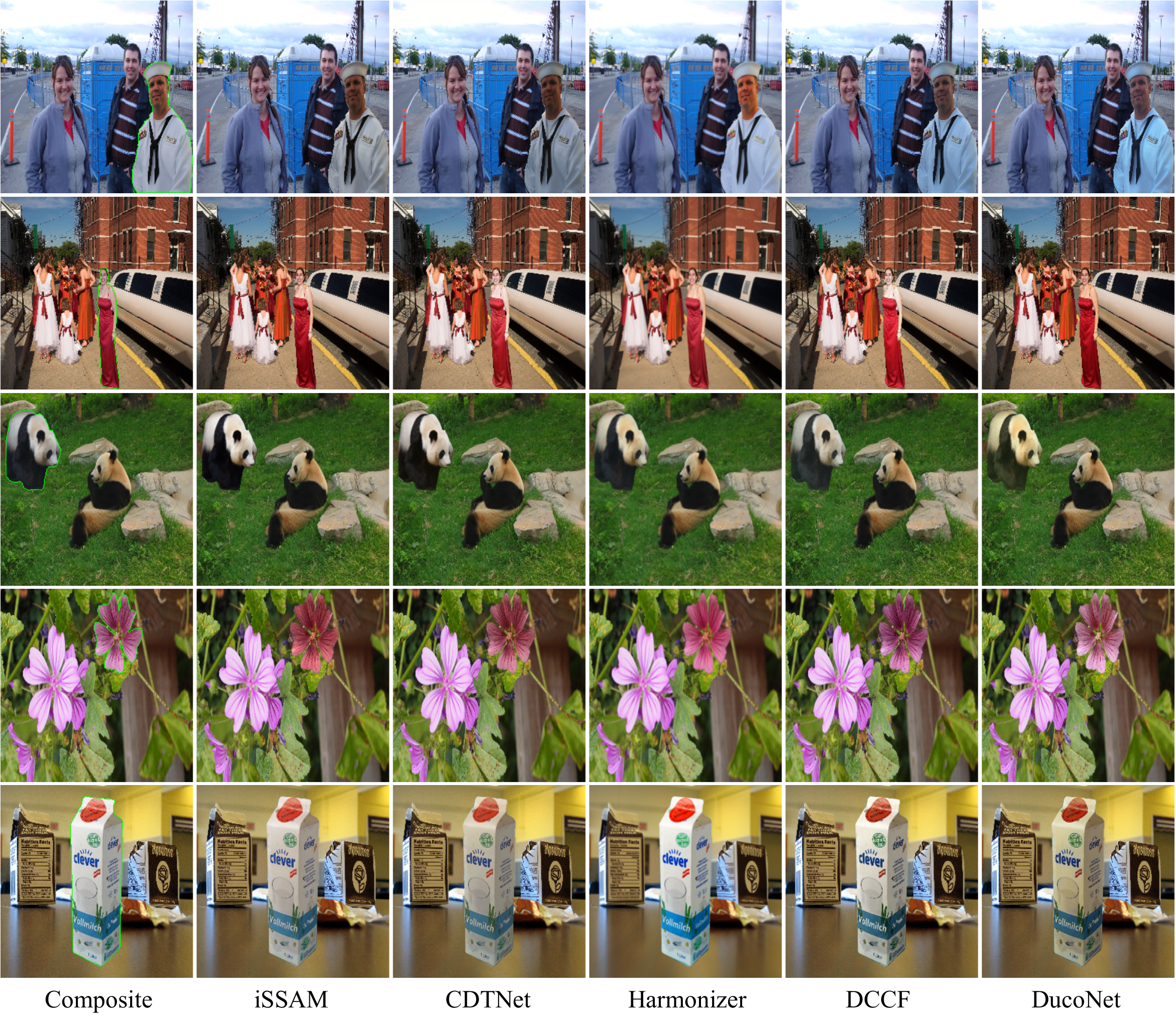}
  \caption{From left to right, we show the composite image ( foreground outlined  in green), the harmonized results of iSSAM~\cite{issam}, CDTNet~\cite{CDTNet}, Harmonizer ~\cite{Harmonizer}, DCCF~\cite{DCCF}, and our DucoNet on real composite images.}
  \Description{}
  \label{fig:hrreal_visual}
\end{figure*}

We further show the visual results of different methods in \Cref{fig:hrreal_visual}.
We observe that our DucoNet produces more harmonious harmonization results.
For example, in rows 1-2, the humans generated by our DucoNet are more consistent with the global context and more harmonious. 
In rows 3, the panda generated by our DucoNet is more consistent with the other panda in the background. 
In rows 4-5, we can generate the objects (\emph{e.g.}, milk carton, flower) that better match the background illuminance.

\begin{table*}[htbp]
    \centering
    \large
    {
    \begin{tabular}{c|c|c|c|c|c|c}
    \toprule
        \textbf{Method}& \textbf{Composite} & \textbf{iSSAM~\cite{issam}} & \textbf{CDTNet~\cite{CDTNet}} & \textbf{Harmonizer~\cite{Harmonizer}} & \textbf{DCCF~\cite{DCCF}} & \textbf{DucoNet}  \\ 
    \midrule
        \textbf{B-T Score} & -1.116 & 0.0781 & 0.176 & 0.130 & 0.272 & 0.460 \\
    \bottomrule
    \end{tabular}}
    \caption{B-T scores of different methods on 100 real composite images \cite{CDTNet}.}
    \label{table:hr100_user_study}
\end{table*}

\begin{figure*}[htbp]
  \centering
  \includegraphics[width=0.95\linewidth]{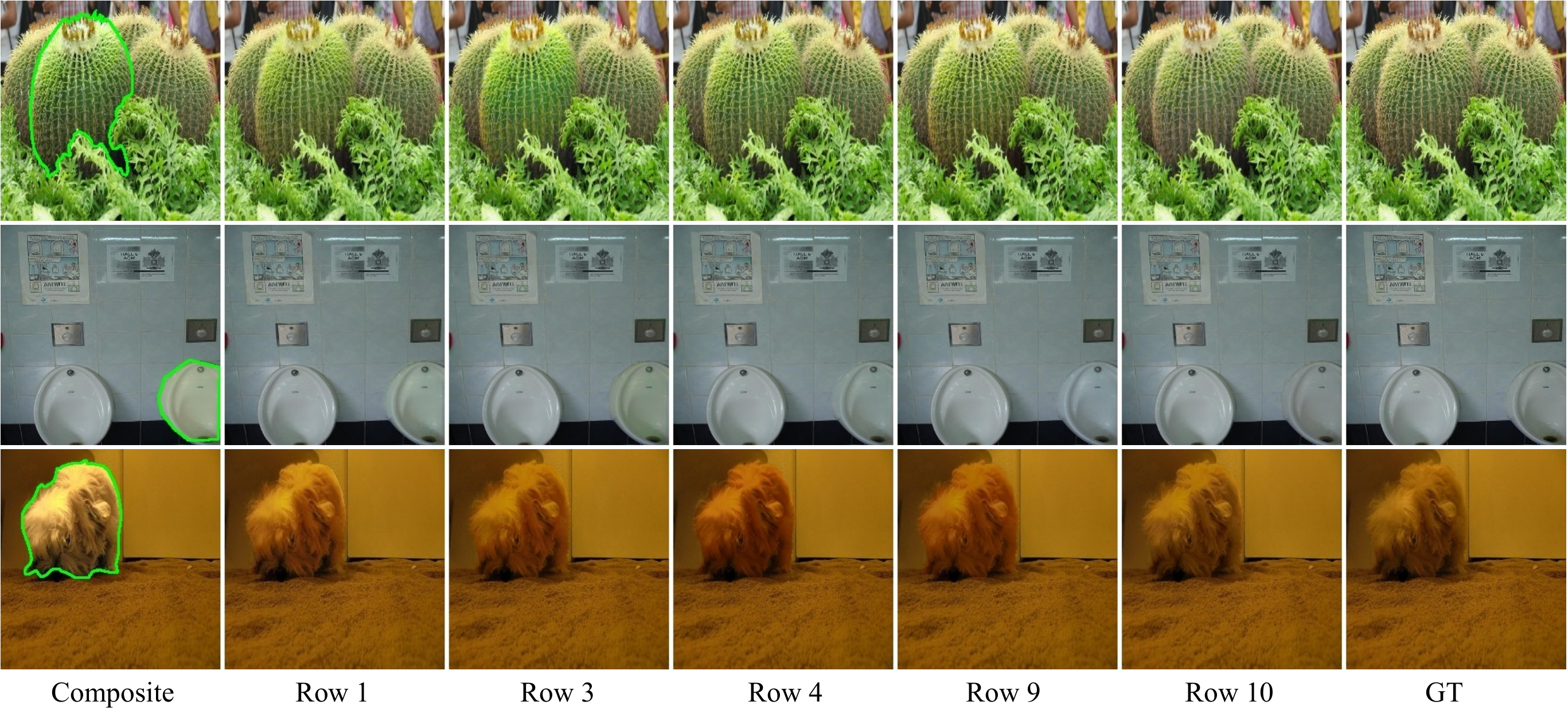}
  \caption{From left to right, we show the composite image ( foreground outlined  in green), the harmonized result of our ablated versions (row 1, 3, 4, 9, 10 in Table 3 of the main submission), and the ground-truth.}
  \Description{}
  \label{fig:ablation_study}
\end{figure*}

\section{Visualization of Ablation Studies}\label{sec:Visualization of Abalation Studies}

In the main submission, we have performed ablation studies of our DucoNet in Section 4.4. 
In \Cref{fig:ablation_study}, we further provide the visualization results of different ablated versions of our DucoNet. 
From left to right, we show the composite image, the harmonized results corresponding to rows 1, 3, 4, 9, and 10 in Table 3 of the main submission, and the ground-truth.
Specifically, row 1 only uses
the input composite image with $RGB$ channels; row 3 (\emph{resp.}, row 4) treats the $Lab$ (\emph{resp.}, $RGB$) channels as a whole input in the encoding module and uses a single control code in the control module; row 9 simply averages three manipulated feature in our $Lab$-CM; and row 10 corresponds to our full DucoNet method.
The visualisation results show that our full DucoNet is closer to the ground-truth and produces more harmonized results.

\begin{figure}[t]
  \centering
  \includegraphics[width=0.95\linewidth]{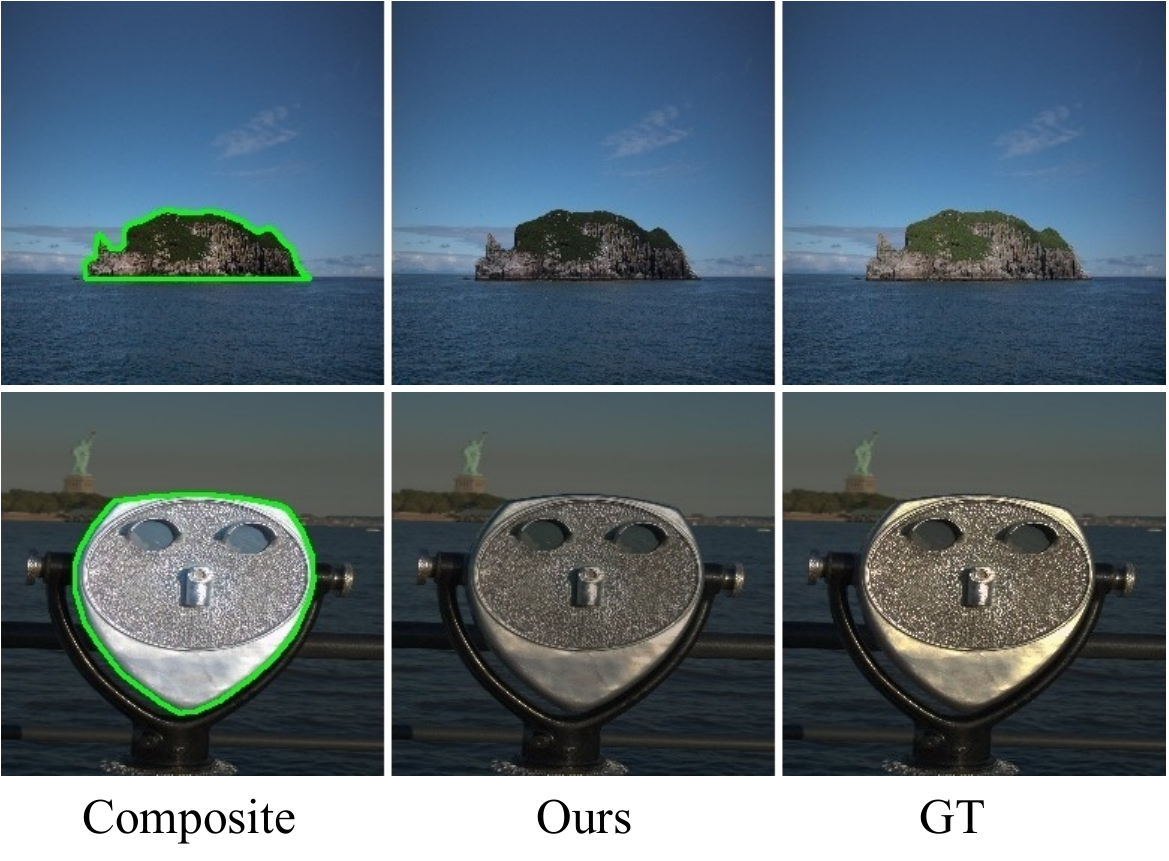}
  \caption{From left to right, we show the composite image (foreground outlined  in green), the harmonized result of our DucoNet, and the ground-truth.}
  \Description{}
  \label{fig:failure_case}
\end{figure}

\section{Failure Cases}\label{sec:Failure Cases}
Although our DucoNet achieves visually harmonious results for most cases, there are still some cases in which our results are inconsistent with the ground-truth as shown in \Cref{fig:failure_case}.
The failure of our DucoNet may be caused by insufficient or incorrect understanding of scene information (\emph{e.g.}, the depth of the foreground and the position of the light source), which remains a challenging task and would lead to unsatisfactory results.

\bibliographystyle{ACM-Reference-Format}
\bibliography{acmart}